# Representing Research Attention as Contextually Structured Flows

Jessica Rodrigues[*], Angelo Salatino[**], Gard Jenset[***], and Scott Hale[*]

[*]*jessica.rodrigues@oii.ox.ac.uk; scott.hale@oii.ox.ac.uk*
University of Oxford, United Kingdom

[**]*angelo.salatino@open.ac.uk*
The Open University, United Kingdom

[***]*gjenset@gmail.com*
Independent researcher, United Kingdom

Research metrics use attention as evidence of societal impact. Yet attention serves as evidence only once interpreted, and its meaning depends on its contextual structure, not on volume alone. Altmetrics represents signals in isolation, keeping a count of the attention an output received, or a sequence of when. We address this with attention flows, representations that situate an output's attention in the contexts through which it is distributed. To evaluate the flow, we build a benchmark of analogy queries, each testing whether the relationship between two outputs, applied to a third, yields a fourth. The count and sequence baselines fail to recover these relationships, whereas flows learned as dynamic contextualised representations recover them. The recovered structure also survives partial observation and rests on its contexts instead of volume. These findings support attention represented as contextually structured for research evaluation.

## 1. Introduction

Societies invest heavily in research, and governments, funders, and institutions increasingly seek evidence that research generates value beyond academia (Bornmann, 2013). Yet many benefits are hard to observe directly, creating a need for evidence that can support interpretations of how research reaches and affects society (Penfield et al., 2014). Increasingly, this role has fallen to *attention*, the interest and engagement that research outputs attract as they are shared, reported, and debated. Attention is observable where impact is not, and has become a principal form of evidence through which societal impact is assessed. But it serves as evidence only once interpreted, and its meaning depends on the contexts in which it occurs, not on volume alone. Altmetrics represents signals in isolation, discarding those contexts. What remains is an impoverished count of how much attention an output received, or a sequence of when.

Two outputs can attract the same volume of attention through entirely different paths. During the COVID-19 pandemic, some outputs drew intense attention immediately across media and policy, often around preliminary findings (Mehra et al., 2020; Gao et al., 2021), while others accumulated it slowly through sustained development feeding into applications such as vaccine design (Pardi et al., 2018). Both reached comparable visibility, yet the contexts through which their attention moved were entirely different. Represented as a count or a sequence, the two are indistinguishable, despite developing in fundamentally different ways. We close this gap by representing attention as a contextually structured object, defined over the contexts through which it is distributed, which we call the *attention flow*.

## 2. Background

Bibliometric and alternative indicators are derived from observable traces of scholarly and societal activity. Citations have long served as proxies for scientific influence (Garfield, 1972), while alternative metrics extended this to online interaction with outputs (Priem et al., 2011; Bornmann, 2013; Sugimoto & Larivière, 2017), with tweets, blogs, and news now read as signals of attention (Haustein et al., 2014). Such signals are not random but fall into recurrent patterns (Wang, Song & Barabási, 2013), where collective attention concentrates and disperses across contexts (Wu & Huberman, 2007; Weng et al., 2012). Research attention is therefore a matter not only of magnitude but of structure, the organisation of attention across the contexts through which it is distributed.

Altmetrics sets this structure aside. Each signal is represented on its own and aggregated into a count, which reports how much attention an output received, or a sequence, which adds when (Priem et al., 2011; Thelwall & Kousha, 2015; Moed, 2005). Neither represents how attention is organised across contexts, nor how that organisation holds together, so distinct developments appear alike once their signals are stripped of context. This is the loss broader critiques of research metrics point to (Wouters et al., 2015; Hicks et al., 2015). From a representation learning perspective, the form of a representation determines which regularities it can capture (Bengio et al., 2013; Goodfellow et al., 2016), which motivates a representation that preserves what altmetrics discards.

## 3. Representing Attention as Flows

A signal's meaning depends on the contexts in which it occurs, much like that of a word (Harris, 1954; Firth, 1957), so a faithful representation must retain those contexts. Attention falls naturally under this account, since the signals through which it is observed are themselves acts of language, each an utterance about a research output. The flow situates an output's signals in the contexts through which its attention is distributed. The contexts are a parameter of the flow, which admits any number of them, so the flow is defined independently of the particular contexts chosen. In this paper we instantiate it over two contexts, each distributed over time. The linguistic context is the language of the signal, how the research output is framed, and the social context is its social setting, the online source in which the signal occurs (Firth, 1950, 1957; Halliday, 1978).

Given this instantiation, we learn an embedding of each flow by applying Dynamic Contextualised Word Embeddings (DCWE; Hofmann et al., 2021) to attention, unchanged, treating each signal as the unit and its output's flow as the target. DCWE combines two axes, contextual and dynamic, giving three variants. Contextualised Word Embeddings (CWE) depend on the linguistic context in which a token appears (Peters et al., 2018; Devlin et al., 2019). Dynamic Word Embeddings (DWE) evolve over time (Bamler & Mandt, 2017). DCWE is both, contextualised and varying with social context and time. Relations between outputs become operations on these flow embeddings (Mikolov et al., 2013; Battaglia et al., 2018), compared against two altmetric baselines, the count and the sequence. Following work that probes what learned representations encode (Hewitt & Manning, 2019; Belinkov, 2022), we probe the flow for three structural properties of attention, invariance, preservation, and dependence, tested next.

## 4. Empirical Evidence of Attention Flows

We construct a dataset from the Scientometric Researcher Access to Data (SRAD) programme, which aggregates publicly observable signals of attention to research outputs across 17 online environments, including social media, news, blogs, patents, policy, etc (Priem et al., 2011; Haustein et al., 2014). From 529,461 signals across 113,789 Computer Science research articles published 2016–2024 (Fields of Research codes 4601–4613), we retain 2,484 outputs with sufficient extent, source diversity, and temporal variation, and split them by output into training and held-out test partitions. From the training partition alone, we split it into train and dev sets and trained flow embeddings across two epochs, yielding three different models: CWE, DWE, and DCWE. From the test partition alone, we build a benchmark of analogy queries of the form A:B::C:D, asking whether the transformation from A to B, applied to C, yields D (Gentner, 1983; Mikolov et al., 2013; Battaglia et al., 2018). Queries are defined against a reference space computed from aggregate features of each output's attention, independent of the representations evaluated, so none is tested against a standard derived from itself. Each output is assigned to one of five regimes, sequential, distributional, adversarial, diffuse, and relational.

*Structural invariance* asks that the structure of attention be general across outputs, so comparable attention exhibits comparable structure. We test this by forming $A - B + C$ for each triplet and retrieving D by cosine similarity, reporting mean percentile rank (MPR), where 0.5 is chance and lower is better (Table 1). If the structure generalises across outputs, a transformation between one pair should transfer to others, placing D near the top of the ranking and MPR below chance. The count and sequence sit at chance, about 2% below, recovering no transferable structure. The flow embeddings fall well below chance, by 7.8% (CWE), 7.6% (DWE), and 8.4% (DCWE), performing best in the adversarial regime, where they reach 32% to 35% below chance. Flows recover comparability that the altmetrics baselines do not, and among them DCWE performs best.

Table 1. Structural invariance: retrieval performance (MPR) across regimes. Lower is better.

| Representation | Sequential | Distributional | Adversarial | Diffuse | Relational | Overall |
|---|---|---|---|---|---|---|
| **Count** | 0.530 | 0.466 | 0.491 | 0.446 | 0.519 | *0.490* |
| **Sequence** | 0.504 | 0.526 | 0.465 | 0.455 | 0.502 | *0.490* |
| **CWE** | 0.517 | 0.483 | 0.338 | 0.451 | 0.514 | *0.461* |
| **DWE** | 0.527 | 0.544 | 0.328 | 0.432 | 0.481 | *0.462* |
| **DCWE** | **0.521** | **0.487** | **0.323** | **0.450** | **0.511** | ***0.458*** |
| **Random** | 0.512 | 0.496 | 0.508 | 0.500 | 0.496 | *0.502* |

*Structural preservation* asks that the structure of attention be stable over time, present as attention unfolds and not only once it is complete. We test this by truncating each output to the earliest 25%, 50%, and 75% of its signals, recomputing the representations, and repeating the retrieval (Table 2). If the structure is stable over time, it should be recoverable from an output's early signals, placing MPR below chance well before its attention is complete. The count stays flat across truncations, holding no structure to recover. The sequence improves only slightly, from 1.4% to 2.4% below chance, but stays near chance throughout. The flow embeddings fall well below chance from only the first quarter, by 5.8% (CWE), 6.0% (DWE), and 6.2% (DCWE), and improve as more is observed, reaching 6.8%, 6.6%, and 7.2% at three quarters. The three variants track closely, with DCWE remaining best across truncations.

Table 2. Structural preservation: retrieval performance (MPR) under truncation across regimes. Lower is better.

| Representation | Regime | Truncation (25%) | Truncation (50%) | Truncation (75%) |
|---|---|---|---|---|

| **Count** | Sequential | 0.530 | 0.530 | 0.530 |
|---|---|---|---|---|
| | Distributional | 0.466 | 0.466 | 0.466 |
| | Adversarial | 0.491 | 0.491 | 0.491 |
| | Diffuse | 0.446 | 0.446 | 0.446 |
| | Relational | 0.519 | 0.519 | 0.519 |
| | Overall | *0.490* | *0.490* | *0.490* |
| **Sequence** | Sequential | 0.507 | 0.501 | 0.497 |
| | Distributional | 0.512 | 0.517 | 0.515 |
| | Adversarial | 0.467 | 0.466 | 0.472 |
| | Diffuse | 0.474 | 0.465 | 0.455 |
| | Relational | 0.505 | 0.501 | 0.501 |
| | *Overall* | *0.493* | *0.490* | *0.488* |
| **CWE** | Sequential | 0.526 | 0.521 | 0.513 |
| | Distributional | 0.484 | 0.484 | 0.488 |
| | Adversarial | 0.345 | 0.332 | 0.332 |
| | Diffuse | 0.477 | 0.468 | 0.475 |
| | Relational | 0.524 | 0.528 | 0.524 |
| | Overall | *0.471* | *0.467* | *0.466* |
| **DWE** | Sequential | 0.538 | 0.531 | 0.531 |
| | Distributional | 0.528 | 0.534 | 0.541 |
| | Adversarial | 0.355 | 0.351 | 0.343 |
| | Diffuse | 0.449 | 0.439 | 0.436 |
| | Relational | 0.479 | 0.481 | 0.486 |
| | Overall | *0.470* | *0.467* | *0.467* |
| **DCWE** | Sequential | 0.530 | 0.525 | 0.519 |
| | Distributional | 0.486 | 0.485 | 0.491 |
| | Adversarial | 0.332 | 0.319 | 0.318 |
| | Diffuse | 0.472 | 0.466 | 0.469 |
| | Relational | 0.522 | 0.524 | 0.522 |
| | Overall | ***0.469*** | ***0.464*** | ***0.464*** |
| **Random** | Overall | *0.512* | *0.512* | *0.512* |

*Structural dependence* asks that the structure of attention rest on its contexts, not on volume alone. We test this by perturbing each output's contexts, reassigning the social setting, language, and timing of its signals across outputs, so each output receives contexts drawn from others while its volume remains unchanged (Table 3). If the structure rests on the contexts, perturbation should degrade retrieval; if it rested on volume, retrieval would be unchanged. The count stays flat, carrying only volume. The sequence degrades by 0.8%, resting only on the thin structure that timing alone holds. The flow embeddings degrade far more, by 5.9% (CWE), 4.1% (DWE), and 5.2% (DCWE). CWE encodes only the linguistic context, while DWE encodes only the social context over time. Since CWE depends more than DWE (5.9% vs 4.1%), and DCWE combines both (5.2%), the linguistic context plays an important role in shaping the structure DCWE encodes. Flows rest on contexts that the altmetrics baselines discard, and the linguistic one weighs most.

Table 3. Structural dependence: retrieval performance (MPR) under perturbation across regimes. Higher Δ is better.

| Representation | Regime | MPR (Original) | MPR (Perturbed) | ΔMPR% |
|---|---|---|---|---|
| **Count** | Sequential | 0.530 | 0.530 | 0.0% |
| | Distributional | 0.466 | 0.466 | 0.0% |
| | Adversarial | 0.491 | 0.491 | 0.0% |
| | Diffuse | 0.446 | 0.446 | 0.0% |
| | Relational | 0.519 | 0.519 | 0.0% |
| | Overall | *0.490* | *0.490* | *0.0%* |
| **Sequence** | Sequential | 0.504 | 0.508 | +0.8% |
| | Distributional | 0.526 | 0.568 | +8.0% |
| | Adversarial | 0.465 | 0.419 | -9.9% |
| | Diffuse | 0.455 | 0.454 | -0.2% |
| | Relational | 0.502 | 0.520 | +3.6% |
| | Overall | *0.490* | *0.494* | *+0.8%* |
| **CWE** | Sequential | 0.518 | 0.567 | +9.5% |
| | Distributional | 0.483 | 0.524 | +8.5% |
| | Adversarial | 0.336 | 0.296 | -11.9% |
| | Diffuse | 0.451 | 0.511 | +13.3% |
| | Relational | 0.514 | 0.536 | +4.3% |
| | Overall | ***0.460*** | ***0.487*** | ***+5.9%*** |
| **DWE** | Sequential | 0.528 | 0.578 | +9.5% |
| | Distributional | 0.542 | 0.574 | +5.9% |
| | Adversarial | 0.327 | 0.288 | -11.9% |
| | Diffuse | 0.429 | 0.468 | +9.1% |
| | Relational | 0.476 | 0.486 | +2.1% |
| | Overall | *0.460* | *0.479* | *+4.1%* |
| **DCWE** | Sequential | 0.520 | 0.569 | +9.4% |
| | Distributional | 0.488 | 0.527 | +8.0% |
| | Adversarial | 0.323 | 0.278 | -13.9% |
| | Diffuse | 0.451 | 0.510 | +13.1% |
| | Relational | 0.512 | 0.533 | +4.1% |
| | Overall | *0.459* | *0.483* | *+5.2%* |
| **Random** | Overall | *0.502* | *0.502* | *0.0%* |

## 5. Conclusion

In this paper, we introduced the attention flow, a representation that situates a research output's attention in the contexts through which it is distributed. We probed it through three structural properties of attention, invariance, preservation, and dependence, and found that the flow captures structure the altmetric baselines do not. Specifically, the structure it recovers is general across outputs, stable as attention develops, and dependent on the contexts rather than volume. This is what lets outputs with comparable volume but different paths be told apart, so attention can be read as evidence of how research reaches society, not only how much. Among the variants, DCWE recovers this structure most fully, making it the most effective way to encode attention flows. Two directions follow, testing flows on downstream impact assessment, and conditioning flows on further contexts. Attention flows are therefore one route toward indicators that preserve what gives attention its meaning.

### Open science practices

This study uses data obtained through the Scientometric Researcher Access to Data (SRAD) programme. While this source provides large-scale coverage of attention signals, they are not fully open, and access is subject to licensing restrictions. For this reason, the raw data cannot be publicly shared. To support transparency, we provide a detailed account of the data collection, filtering, and preprocessing pipeline.

All derived artefacts are made openly available. This includes the benchmark, as well as the source code available at https://github.com/jehrodrigues/representing-research-attention-as-contextually-structured-flows. No formal preregistration was conducted, as the study is exploratory, but all methodological choices are explicitly specified.

### Acknowledgments

The authors thank Digital Science for access to the data used in this study, provided through the SRAD programme.

### Author contributions

The first author led the study, including the attention flow representation, research design, experiments, and manuscript. The remaining authors supervised and advised on framing, modelling, evaluation, and scientometric context.

### Competing interests

The authors declare that they have no competing interests.